%% file: 0.main.tex
\documentclass[11pt]{article}

\usepackage[preprint]{acl}

\usepackage{subcaption}

\usepackage{algorithm}
\usepackage[noend]{algpseudocode}

\usepackage{times}
\usepackage{latexsym}

\usepackage[T1]{fontenc}
\usepackage[utf8]{inputenc}

\usepackage{microtype}
\usepackage{inconsolata}
\usepackage{soul}

\usepackage{graphicx}

\usepackage{booktabs}
\usepackage{float}

\usepackage{tikz}
\usetikzlibrary{shapes.multipart, positioning, decorations.markings, arrows.meta}
\tikzset{
    gm_var/.style={
        circle,
        draw,
        fill=white,
        inner sep=.25em
    },
    gm_factor/.style={
        rectangle,
        fill=black,
        inner sep=.25em
    }
}
\usepackage{amsmath}

\usepackage{amsthm}

\usepackage{mdframed}
\definecolor{theoremcolor}{rgb}{0.95, 0.95, 0.95}
\mdfsetup{
    backgroundcolor=theoremcolor
}
\theoremstyle{definition}
\newmdtheoremenv{thm}{Proposition}
\newmdtheoremenv{definition}{Definition}

\usepackage{enumitem}
\usepackage{amssymb}

\usepackage{cancel}
\usepackage{bm}
\usepackage{wasysym}
\input{math_commands}

\def\vlambda{{\bm{\lambda}}}
\def\vnu{{\bm{\nu}}}
\usepackage[only,llbracket,rrbracket]{stmaryrd}
\newcommand\maxOmega{\textstyle{\max_\Omega}}
\DeclareMathOperator{\conv}{conv}
\DeclareMathOperator{\dom}{dom}

\allowdisplaybreaks

\usepackage[textsize=tiny]{todonotes}

\newcommand\MRF{\textsc{Mrf}}
\newcommand\CRF{\textsc{Crf}}
\newcommand\MF{\textsc{Mf}}
\newcommand\BCRF{\textsc{Bcrf}}

\usepackage{xspace}
\makeatletter
\DeclareRobustCommand\onedot{\futurelet\@let@token\@onedot}
\def\@onedot{\ifx\@let@token.\else.\null\fi\xspace}

\def\eg{\emph{e.g}\onedot} 
\def\ie{\emph{i.e}\onedot}

\makeatother

\title{Bregman Conditional Random Fields: Sequence Labeling\\
with Parallelizable Inference Algorithms}

\author{
  \textbf{Caio Corro\textsuperscript{1}}\qquad
  \textbf{Mathieu Lacroix\textsuperscript{2}}\qquad
  \textbf{Joseph Le Roux\textsuperscript{2}}
\\
  \textsuperscript{1}INSA Rennes, IRISA, Inria, CNRS, Université de Rennes, France
\\
    \textsuperscript{2}Université Sorbonne Paris Nord, CNRS, LIPN, France
\\
  \texttt{caio.corro@irisa.fr}\qquad
  \texttt{\{lacroix,leroux\}@lipn.fr}
}

\begin{document}
\maketitle
\begin{abstract}
We propose a novel discriminative model for sequence labeling called Bregman conditional random fields (\BCRF{}).
Contrary to standard linear-chain conditional random fields,
\BCRF{} allows fast parallelizable inference algorithms based on iterative Bregman projections.
We show how such models can be learned using Fenchel-Young losses, including extension for learning from partial labels.
Experimentally, our approach delivers comparable results to \CRF{} while being faster, and achieves better results in highly constrained settings compared to mean field, another parallelizable alternative.
\end{abstract}

\input{1.introduction}
\input{2.viterbi}
\input{3.proposed_algorithm}

\input{4.training}
\input{5.related_work}

\input{6.experiments}
\input{7.conclusion}

\section*{Acknowledgements}

This work was granted access to the HPC resources of IDRIS under the allocation 2024-AD011013727R1 made by GENCI.
This work was supported by the LABEX EFL (Empirical Foundations of Linguistics, ANR-10-LABX-0083), operated by the French National Research Agency
(ANR).
This work is supported by the SEMIAMOR (CE23-2023-0005) and InExtenso (ANR-23-IAS1-0004) project grants given by
the French National Research Agency (ANR).

\input{8.limitations}

\bibliography{custom}

\appendix
\input{9.dp}
\input{9.logsumexp}
\input{9.subgradient_fenchel_conjugate}
\input{9.proof_entropic_reg}
\input{9.proof_polytope}

\input{9.lightspeed}
\input{9.mft}
\input{9.loss}

\input{9.exp}

\end{document}

%% file: math_commands.tex
\def\1{\bm{1}}

\def\rv{{\textnormal{v}}}

\def\vzero{{\bm{0}}}
\def\vone{{\bm{1}}}
\def\vmu{{\bm{\mu}}}
\def\vtheta{{\bm{\theta}}}

\def\vp{{\bm{p}}}
\def\vq{{\bm{q}}}

\def\vs{{\bm{s}}}
\def\vt{{\bm{t}}}
\def\vu{{\bm{u}}}
\def\vv{{\bm{v}}}
\def\vw{{\bm{w}}}
\def\vx{{\bm{x}}}
\def\vy{{\bm{y}}}

\def\evq{{q}}

\def\evs{{s}}

\def\evu{{u}}
\def\evv{{v}}
\def\evw{{w}}
\def\evx{{x}}
\def\evy{{y}}

\def\mA{{\bm{A}}}
\def\mB{{\bm{B}}}

\def\mM{{\bm{M}}}

\DeclareMathAlphabet{\mathsfit}{\encodingdefault}{\sfdefault}{m}{sl}
\SetMathAlphabet{\mathsfit}{bold}{\encodingdefault}{\sfdefault}{bx}{n}

\def\emA{{A}}
\def\emB{{B}}

\newcommand{\E}{\mathbb{E}}

\newcommand{\R}{\mathbb{R}}

\newcommand{\KL}{D_{\mathrm{KL}}}

\DeclareMathOperator*{\argmax}{argmax}
\DeclareMathOperator*{\argmin}{argmin}

%% file: 1.introduction.tex
\section{Introduction} \label{sec:introduction}

Sequence labeling is a core natural language processing task:
it consists in assigning one tag per token of an input sentence.
Although simple, it is powerful enough to encompass part-of-speech (POS) tagging~\cite{church-1988-stochastic},
named-entity recognition~\cite{ramshaw-marcus-1995-text} including complex variants \cite{corro2024discner},
word segmentation~\cite{xue-2003-chinese},
and even syntactic and semantic parsing~\cite{xipeng-2009,marcheggiani-etal-2017-simple,gomez2018sequencelabeling}.

The main structured models for sequence labeling are hidden Markov models \cite{jelinek-1997-statis-method} and conditional random fields \cite[\CRF{},][]{lafferty-2001-condit-random-field}.
In both settings, the most probable sequence of tags is computed via the Viterbi algorithm~\cite{viterbi1967,forney1973viterbi}.
Learning parameters via maximum likelihood requires to compute the log-partition function of the exponential family distribution over all possible tag sequences \cite{wainwright2008expfam}, which can be done using the Forward algorithm \cite{rabiner-1989-tutor-hidden}.

Both algorithms have a linear-time complexity in the input length.
They are naturally expressed as dynamic programming algorithms that recursively solve subproblems.
Unfortunately, these algorithms are inherently sequential and therefore cannot benefit from parallelization of modern GPUs.

We propose a radically different approach named Bregman conditional random fields (\BCRF), inspired by entropic optimal transport \cite{cuturi2013lightspeed,benamou-2015-iterat-bregm} and \textsc{SparseMAP} \cite{niculae-2018-spars}.
\BCRF{} relies on mean regularization~\cite{blondel2020fy} to define probability distributions over sequence labelings.
In this setting, we develop approximate inference algorithms based on iterative Bregman projections \cite{bregman-1967},
where each step of the algorithm solves parallelizable  subproblems, taking full advantage of modern hardware \cite{censor1998ibp}.
Empirically we show that this method is accurate and compares favorably to alternatives like mean field \cite[\MF,][]{wang-etal-2020-ain}, especially in the presence of forbidden tag transitions, a typical requirement in segmentation and named entity recognition.

Our contributions can be summarized as follows:
(1)~we introduce \BCRF{}, a novel approach to define distributions over sequence labelings;
(2)~we develop a strongly parallelizable inference algorithm that can be used as a drop-in replacement for both Viterbi and Forward;
(3)~we show how Fenchel-Young losses \cite{blondel2020fy} can be leveraged to learn the parameters of a \BCRF{} model, including the case of learning from partial labels;
(4)~we experiment on POS tagging, token segmentation and named entity recognition and show that our approach delivers the same performance as standard \CRF{} while being faster.

Our code is publicly available.\footnote{\url{https://github.com/FilippoC/lightspeed-crf}}

%% file: 2.viterbi.tex
\section{Background}\label{sec:background}

In this section, we set notations and review \CRF{} to stress what the computational limitations are and how our approach differs from standard \CRF.

\paragraph{Notations.}
We write $\llbracket a, b \rrbracket$ the set of integers from $a$ to $b$.
Given a discrete set $S$, we write $\R^S$ the vector of reals indexed by elements of $S$, and similarly for higher dimension tensors.
Given set $S$ and function $h$, we denote $\vv = \left[ h(s) \right]_{s \in S}$ the vector indexed by elements of $S$ s.t.\ $\evv_s = h(s)$.
Given matrices $\mA$ and $\mB$, we denote $\langle \mA, \mB\rangle = \sum_{i, j} \emA_{ij}\emB_{ij}$ the sum of entries of the Hadamard product (\ie{} dot product for vectors).
The $d\!-\!1$ dimension simplex is denoted $\triangle(d) =\{\vx\in\mathbb{R}_{+}^{d} : \|\vx\|_{1} = 1 \}$, and we write $\triangle$ when dimension can be inferred from context and $\triangle(S)$ when dimensions are indexed by elements of $S$.
Given vectors $\vv \in \R^d_{+}$ and $\vt \in \R^d_{++}$,
the Shannon entropy and the Kullback-Leibler divergence\footnote{Although the definition of $\KL$ may seem surprising, it is the specific form used for iterative Bregman projections as it corresponds to a Bregman divergence \cite{benamou-2015-iterat-bregm}.} are respectively defined as:
\begin{align*}
    H(\vv)
    &= - \langle \vv, \log \vv \rangle,
    \\
    \KL[\vv | \vt]
    &= \left\langle \vv, \log \frac{\vv}{\vt}\right\rangle - \langle \vv, \vone\rangle.
\end{align*}
Given a convex set $S$,
we denote $\delta_S$ the indicator function of $S$:
\begin{align*}
    \delta_S(\vs)
    = \begin{cases}
        0\quad&\text{if~} \vs \in S, \\
        \infty\quad&\text{otherwise.}
    \end{cases}
\end{align*}
Given function $h$, we note $h^*$ its Fenchel conjugate:
\[
h^*(\vv) = \sup_{\vmu \in \dom h} \langle \vv, \vmu\rangle - h(\vmu)\,, 
\]
and use $\max$ instead of $\sup$ when maximum exists.

\subsection{Conditional Random Fields}

Without loss of generality, we assume all inputs of length $n \ge 3$.
Let $\vs = \evs_1 \dots \evs_n$ be an input sentence and $T$ the set of tags.
We denote by $X$ the set of sentence labelings (or tag sequences) for \(\vs\), \ie{} $\vx = \evx_1 \dots \evx_n \in X$ where $\evx_i \in T$ is the tag associated with word $s_{i}$.

A Markov random field \cite[\MRF,][]{hammersley1971mrf} defines a distribution over elements of $X$ using a collection of sufficient statistics $F$:
\[
    \phi(\vx) \triangleq \left( \phi_f(\vx) \right)_{f \in F}.
\]
For sequence tagging, it is usual to rely on a first-order linear-chain, a \MRF{} composed of unary and sequential binary sufficient statistics.
Unary sufficient statistics are one-hot vectors representing each tag $\evx_{i}$,
while binary sufficient statistics indicate transitions between consecutive tags:
\begin{align*}
\mathsf{bi}_{i}(\vx)
&= \underbrace{[0 \dots 1 \dots 0]}_{\text{$|T|\times|T|$ values}}\,,
\end{align*}
\ie{}\ $\mathsf{bi}_{i}(\vx)_{t,t'}=1$ if $\evx_{i}=t$ and $\evx_{i+1}=t'$, and 0 otherwise.
Note that in the case of linear chains, unary sufficient statistics are redundant.
Therefore, in the following we will use only binary sufficient statistics and denote sufficient statistics as vectors:
\[
  \phi(\vx) 
  \in \{0,1\}^W
\]
where $W = \llbracket 1, n-1 \rrbracket \times T \times T$, and 
$\phi(\vx)_{i, t, t'} = 1 \Leftrightarrow \evx_i = t \land \evx_{i+1} = t'$.
We denote the set of sufficient statistics as
$
Y = \{ \phi(\vx) | \vx \in X \}
$.

The exponential family associated with $\phi$ is:
\[
p_\vw(\vx) = \exp\left(~\langle \vw, \phi(\vx)\rangle - A_Y(\vw)~\right),
\]
where $\vw \in \R^W$ is the vector of canonical parameters weighting
statistics.
More precisely $\evw_{i, t, t'}$ is the transition score associated with tagging
$s_{i}$ with $t$ and $s_{i+1}$ with $t'$, reflected by the sufficient statistics $\mathsf{bi}_{i}(\vx)_{t,t'}$.
To forbid tag assignments, or transitions,
one can set the corresponding values in $\vw$ to $-\infty$.
$A_Y$ is the log-partition function ensuring that the distribution is correctly normalized:
\begin{align}
\label{eq:partition}
A_Y(\vw)
&= \log \sum_{\vy \in Y} \exp~\langle \vw, \vy \rangle.
\end{align}
Computing the most probable tag sequence
$\widehat{\vx}(\vw)$
can be reduced to solving:
\begin{align}
    \label{eq:map}
    \widehat{\vx}(\vw)
    &\in \argmax_{\vx \in X}~\langle \vw, \phi(\vx) \rangle.
\end{align}

A conditional random field \cite[\CRF,][]{lafferty-2001-condit-random-field}
is a \MRF{} for which the canonical parameters $\vw$ are conditioned on an external observation, the input sentence $\vs$ in our case.
To this end, their values are computed by a $\theta$-parameterized scoring function $f_\theta$,  \eg{} a neural net.
Then, the conditional probability of $\vx$ given $\vs$ can be written as:
\[
p_\theta(\vx | \vs)
=
\exp\left(~\langle f_\theta(\vs), \phi(\vx)\rangle - A_Y(f_\theta(\vs))~\right).
\]
Parameters $\theta$ are usually learned from a labeled dataset $(\vs, \vx) \in D$  by minimizing the expected negative log-likelihood (NLL) of the data:
\[
\min_\theta~\E_{(\vs, \vx) \sim D}\left[~\ell_\text{NLL}(f_\theta(\vs) ; \phi(\vx))~\right],
\]
where the NLL loss function is defined as:
\[
\ell_{\text{NLL}}(\vw ; \vy) = - \langle \vw, \vy \rangle + A_Y(\vw)\,.
\]

\subsection{The Viterbi and Forward Algorithms \label{sec:viterbi}}

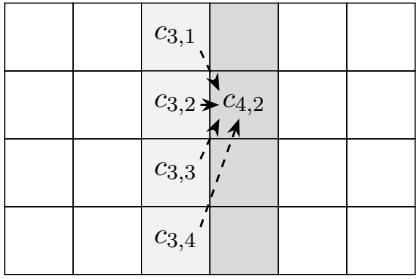
\begin{figure}[h]
\centering
\input{figures/viterbi}
\caption{Illustration of the dependency between stages in the dynamic programming recursion.}
\label{fig:stages}
\end{figure}

We now give a brief presentation of the Viterbi and forward algorithms to compute \eqref{eq:map} and \eqref{eq:partition}, respectively, using the framework proposed by \citet{mensch2018diffdp}.
For any vector $\vv \in \R^d$, we define the $\Omega$-regularized maximum function \cite{nesterov2004smooth,niculae2017regularized}:
\begin{align*}
    \maxOmega(\vv)
    =
    \max\limits_{\vmu \in \triangle(d)}~\langle \vv, \vmu  \rangle - \Omega(\vmu).
\end{align*}
In particular, we can recover two known functions.
First, the \emph{hard} maximum via null regularization:
\begin{align*}
  \Omega(\cdot) = 0 ~&\Leftrightarrow~\maxOmega(\vv) = \max_{i}~\evv_i.
\end{align*}
Second, we can express the \emph{soft} maximum (also called \emph{logsumexp}) via entropic regularization:
\begin{align}
\label{eq:logsumexp}
    \Omega = -H &\Leftrightarrow~\maxOmega(\vv) = \log \sum_i \exp \evv_i\,,
\end{align}
see Appendix~\ref{app:logsumexp}.

Let $c_\Omega(\vw)$ be the regularized maximum score over tag sequences $X$:
\[
c_\Omega(\vw) \triangleq \maxOmega\left(~\left[ \langle \vw, \phi(\vx) \rangle \right]_{\vx \in X}~\right)\,.
\]
We can recast computing equations \eqref{eq:map} and \eqref{eq:partition} as:
\begin{align*}
    c_0(\vw)
    = \max_{\vx \in X}~\langle \vw, \phi(\vx)\rangle
    ~\text{and}~
    c_{-H}(\vw)
    = A_Y(\vw),
\end{align*}
\noindent highlighting the fact that MAP and marginal inference only differ in the choice of regularization.

Although $|X|$ is exponential in the length of the input sentence, we can decompose the computation of $c_\Omega(\vw)$ into a sequence of subproblems, \ie{}~we can recursively compute intermediate chart values $c_{\Omega, i, t}$ via dynamic programming \cite{bellman1954dp} :
\begin{align*}
c_{\Omega, 1, t}(\vw) &\triangleq 0, \\
c_{\Omega, i,t}(\vw) &\triangleq \maxOmega\left(~\left[c_{\Omega, i-1, t'}(\vw) + \evw_{i, t', t}\right]_{t' \in T}~\right), \\
c_\Omega(\vw) &\triangleq \maxOmega\left(~\left[c_{\Omega, n, t}(\vw) \right]_{t \in T}~\right).
\end{align*}
Proof is given in Appendix~\ref{app:dp}.
Space and time complexities are $\mathcal O(n|T|)$ and \(\mathcal O(n|T|^2)\), respectively.

\paragraph{Wavefront Parallelization.}
For a position $i$, chart values $c_{\Omega, i, t}$  do not depend on each other.
We can group chart elements in stages corresponding to each position $i$.
Elements in stage $i+1$ directly depend on elements in stage $i$, see Figure~\ref{fig:stages}.
Most implementations of the Viterbi and forward parallelize computations in a given stage, a technique called \emph{wavefront parallelization} \cite{muraoka1971parallelism}.
Unfortunately, due to the dependency between stages, the $n$ parallel computations must be performed sequentially, which prevent leveraging parallelization capabilities of GPUs.

%% file: figures/viterbi.tex
\begin{tikzpicture}
  \def\cellsize{0.9}

  \foreach \row in {0,1,2,3} {
    \foreach \col in {0,1,2,3,4,5} {
      \ifnum\col=2
        \fill[gray!10] (\col*\cellsize,-\row*\cellsize) rectangle ++(\cellsize,-\cellsize);
      \fi
      \ifnum\col=3
        \fill[gray!30] (\col*\cellsize,-\row*\cellsize) rectangle ++(\cellsize,-\cellsize);
      \fi
      \draw (\col*\cellsize,-\row*\cellsize) rectangle ++(\cellsize,-\cellsize);
    }
  }

  \node[inner sep=1pt] (s1) at ({2*\cellsize + 0.5*\cellsize}, {-0*\cellsize - 0.5*\cellsize}) {$c_{3,1}$};
  \node[inner sep=1pt] (s2) at ({2*\cellsize + 0.5*\cellsize}, {-1*\cellsize - 0.5*\cellsize}) {$c_{3,2}$};
  \node[inner sep=1pt] (s3) at ({2*\cellsize + 0.5*\cellsize}, {-2*\cellsize - 0.5*\cellsize}) {$c_{3,3}$};
  \node[inner sep=1pt] (s4) at ({2*\cellsize + 0.5*\cellsize}, {-3*\cellsize - 0.5*\cellsize}) {$c_{3,4}$};
  
  \node[inner sep=1pt] (t) at ({3*\cellsize + 0.5*\cellsize}, {-1*\cellsize - 0.5*\cellsize}) {$c_{4,2}$};

  \draw[dashed,->,thick,>={Stealth[scale=1]}] (s1.south east) -- (t.north west);
  \draw[dashed,->,thick,>={Stealth[scale=1]}] (s2) -- (t);
  \draw[dashed,->,thick,>={Stealth[scale=1]}] (s3.north east) -- (t.south west);
  \draw[dashed,->,thick,>={Stealth[scale=1]}] (s4.north east) -- (t);
  
\end{tikzpicture}

%% file: 3.proposed_algorithm.tex
\section{Bregman Conditional Random Fields}\label{sec:bregman}

We now introduce \BCRF{} and link maximum a posteriori (MAP) inference with marginal inference.
Then, we show how marginal inference for \BCRF{} can be reduced to a Kullback-Liebler projection into the intersection of two convex sets.
This paves the way for inference via iterative Bregman projections.

\subsection{Mean Regularization}
\label{sec:meanReg}

Let $\mM = \{0, 1\}^{W \times Y}$ be the matrix whose columns are the vectors of $Y$, \ie{}\ $\left[\mM^\top\right]_\vy = \vy$.
As such, $\mM^\top \vw$ is a vector containing the score of each valid tag sequence.
From Equation~\eqref{eq:logsumexp},
we rewrite the log-partition function as:
\begin{align}
    A_Y(\vw)
    \label{eq:distrib_reg}
    &=
    \max_{\vp \in \triangle(Y)}~
        \left\langle
            \vp,
            \mM^\top \vw
        \right\rangle
        + \underbrace{H(\vp)}_{\text{distrib.\ reg.}},
    \\
\intertext{where the entropy regularizes the distribution over all outputs.
Setting $\vq = \mM \vp$ and optimizing over $\vq \in \{ \mM \vp | \vp \in \triangle(Y)\} = \conv(Y)$, we obtain:}
\label{eq:structured_reg}
  A_Y(\vw)  &=
    \max_{\vq \in \conv(Y)}~
    \langle \vq, \vw \rangle - R(\vq)\,,
\end{align}
where $R$ is a structured regularization term, defined so that equality holds.\footnote{See \cite[][Eq.~25]{blondel2020fy} for the exact definition.}
The set $\conv(Y)$ is called the marginal polytope, and computing $A_Y(\vw)$ is therefore often referred to as marginal inference.
The gradient of $A_Y$ is defined as (Appendix \ref{app:subgradient}):
\begin{align}
    \nabla A_Y(\vw)
    &= \argmax_{\vq \in \conv(Y)}~\langle \vq, \vw \rangle - R(\vq)\,,
    \label{eq:structured_reg_grad}
\end{align}
\ie{}\ $\nabla A_Y(\vw)$ is a vector of marginal probabilities, and there exists a unique distribution over sequences $\vp$ such that $\nabla A_Y(\vw) = \mM \vp$.
As such, function $A_Y(\vw)$ identifies an exponential family distribution over sequences $X$.

It is important to note that the dimension of $\vq \in \conv(Y)$ is polynomial in the input length, while the dimension of $\vp \in \triangle(Y)$ is exponential in the input length.
Key to our approach is to use another family of distributions over sequences where regularization is applied on marginals, called mean regularization.
\begin{definition}
Given cannonical parameters $\vw \in \R^W$,
a \BCRF{} defines a distribution over sequence labeling $Y$ whose marginal transition probabilities are given by $\nabla B_Y(\vw)$ where:
\begin{align*}
    B_Y(\vw)
    &=
    \max_{\vp \in \triangle(Y)}
    \langle \vp, \mM^\top \vw \rangle + \underbrace{H(\mM \vp)}_{\text{mean reg.}}.
\end{align*}
\end{definition}
Using the same change of variable, we obtain:
\begin{align*}
     B_Y(\vw) &=
    \max_{\vq \in \conv(Y)}
    \langle \vq, \vw \rangle + H(\vq).
\end{align*}
As in Equation~\eqref{eq:structured_reg}, the optimization is done over vector $\conv(Y)$
but the regularization has now a simple analytical form, paving the way for efficient solvers.

\paragraph{MAP inference.}
Computing the most probable $\vy \in Y$ reduces to solving an unregularized problem for both distributions defined by $A_Y$ and $B_Y$:
\begin{align*}
    \widehat{\vy}(\vw) &\triangleq
    \argmax_{\vy \in \conv(Y)}~\langle \vw , \vy \rangle,
\intertext{and the best tag assignation is then:}
\widehat{\vx}(\vw)_{i}
&=
\argmax_{t \in T}~\max_{t' \in T}~\widehat{\vy}(\vw)_{i, t, t'}
\\
\widehat{\vx}(\vw)_{n}
&=
\argmax_{t \in T}~\max_{t' \in T}~\widehat{\vy}(\vw)_{n-1, t', t}
\end{align*}
for $i \in \llbracket 1, n\!-\!1\rrbracket$.
Although seemingly different from marginal inference, the next proposition shows that this problem can be approximated by:
\begin{align*}
    \nabla B_Y(\tau^{-1}\vw) &=
    \argmax_{\vq \in \conv(Y)}~\langle \tau^{-1}\vw , \vq \rangle +  H(\vy),
\end{align*}
for a sufficiently small temperature $\tau > 0$.

\begin{thm}\label{th:limit}
\[
\lim_{\tau \to 0} B_Y(\tau^{-1}\vw)
=
\max_{\vy \in \conv(Y)}~\langle \vw, \vy\rangle.
\]    
\end{thm}
The proof is given in Appendix~\ref{app:proof_limit}.
Although Proposition~\ref{th:limit} tells that mean regularization can be used as approximate MAP inference,
setting \(\tau\) close to zero raises several issues:
(1)~the algorithm we develop in Section~\ref{sec:ibp} cannot be applied when $\tau = 0$, and (2)~too small values for $\tau$ may lead to computational instabilities such as over- and underflow.
Therefore, we fix $\tau$ to a small value and then approximate MAP by searching for the most probable tag after marginalizing over transition probabilities:
\begin{align*}
\widehat{\vx}(\vw)_{i}
&\simeq 
\argmax_{t \in T}
\sum_{t' \in T} \left[ \nabla B_Y(\tau^{-1} \vw) \right]_{1, t, t'}
\\
\text{and~~}
\widehat{\vx}(\vw)_{n}
&\simeq 
\argmax_{t \in T}
\sum_{t' \in T} \left[ \nabla B_Y(\tau^{-1} \vw) \right]_{n-1, t', t}
\end{align*}
for $i \in \llbracket 1, n\!-\!1 \rrbracket$.
This can be understood as minimum Bayes risk decoding \cite[MBR,][]{bickel2015,goodman1996mbr},
with risk defined as the number of incorrect predictions.\footnote{As such, MBR decoding can return an invalid sequence of tags. In the seminal work of \citet{goodman1996mbr} for syntactic parsing, this was motivated by the fact that evaluation metric did not take into account the well-formedness of predictions.}

\subsection{Marginal Polytope}
\label{sec:reformulation}

The marginal polytope $\conv(Y)$ plays an important role in \BCRF{} as inference requires to optimize a concave objective over it.
In this section, we give a novel tight characterization of this polytope, that will be used to develop our algorithms.
Key is our focus on transitions instead of tags.

\paragraph{Reduction.}
We formalize interpreting each vector in $Y$ as a set of arc selection variables in a graph.
Let $G=(V,E)$ be a directed graph.
The node set $V$ is partitioned into clusters $V_i = \{\rv_i^t\mid t \in  T\}$ associated with each word $s_i$, for $i \in \llbracket 1,n \rrbracket$.
The arc set $E$ consists of arcs connecting nodes of a cluster to nodes of the consecutive one:
\[
E = \{(\rv_i^t, \rv_{i+1}^{t'})\mid i \in \llbracket 1,n-1 \rrbracket, t \in T, t' \in T\}.
\]
By construction, any path from a node of $V_1$ to one of $V_n$ covers a node in each cluster, and covering node $\rv_i^t$ corresponds to assigning tag $t$ to word $s_i$. Such a path is called \emph{valid}.
It is is easy to see that there is one-to-one mapping between valid paths and elements of Y:
each vector $\vy \in Y$ corresponds to the valid path than contains arc $(\rv_i^t, \rv_{i+1}^{t'})$ if and only if $\vy_{i, t, t'} = 1$.

\paragraph{Linear Programming.}
We now give a description of the convex hull of $Y$.
For a node subset $U \subseteq V$, let $\delta^+(U)$ and $\delta^-(U)$ denote the set of arcs leaving and entering $U$, respectively. For simplicity, we write $\delta^+(\rv)$ and $\delta^-(\rv)$ instead of $\delta^+(\{\rv\})$ and $\delta^-(\{\rv\})$.

Any valid path enters each cluster but $V_1$ once, which can be expressed as:\footnote{Since $G$ is acyclic and arcs connect consecutive clusters, only one of these constraints is necessary, the others becoming redundant.
However, they are kept in order to strengthen each subproblem in our decomposition.}
\begin{equation}
\forall i \in \llbracket 2,n-1 \rrbracket: \sum_{a \in \delta^-(V_i)} \evy_a = 1.
\label{cst:enteringCluster}
\end{equation}
Next, for any node not in $V_1$ nor $V_n$, the number of incoming arcs must be egal to the number of outgoing arcs (which will be either 0 or 1 in an integral solution):
\begin{align}
    \begin{array}{l}\forall i \in \llbracket 2, n-1 \rrbracket,\\ \forall \rv \in V_i\end{array}:
    \sum_{a \in \delta^+(\rv)} \evy_a = \sum_{a \in \delta^-(\rv)} \evy_a.
    \label{cst:flow}
\end{align}
It is easy to see that any valid path in $G$ satisfies these constraints. Since $G$ is acyclic, any vector $\vy \in \{0, 1\}^E$ that satisfies equalities \eqref{cst:enteringCluster} and \eqref{cst:flow} describes a valid path.
The next proposition shows that they define an integral polytope, \ie{} they characterize the convex hull of feasible solutions.

\begin{thm}\label{thm:integrality}
The tagging polytope $\conv(Y)$ is described by:
\[
\conv(Y) = \{ \vy \in \mathbb{R}^E_+ \mid \vy \text{ satisfies  \eqref{cst:enteringCluster}, \eqref{cst:flow}}.
\}\]    
\end{thm}
The proof is given in Appendix~\ref{app:proof_marginal_polytope}.

\subsection{Iterative Bregman Projections}
\label{sec:ibp}

\begin{algorithm*}
\begin{algorithmic}
\small
\Function{BcrfInference}{$\vw, n, k$}
    \State{$\vq^{(0)} \gets \vw$}
    \For{$i \gets 1 \dots k$} 
        \State $\vq^{(i)} \in \R^W$
        \Comment{Randomly initialized vector}
        \If{$i \equiv 0 \pmod 2$}
            \For{$j \gets 1 \dots \lceil \frac{n}{2} \rceil - 1$}
            \Comment{Can be computed in parallel}
                \State
                $
                    \vq^{(i)}\vert_{\delta(V_{2j})}
                    \gets
                    \argmin\limits_{
                        \vmu \in \R^W
                        ~\text{s.t.}~
                        \vmu\vert_{\delta(V_{2j})}
                        \in \mathcal{C}_{2j}
                    }
                    \KL\left(
                        \vmu\vert_{\delta(V_{2j})},
                        \vq^{(i-1)}\vert_{\delta(V_{2j})} 
                    \right)
                $
            \EndFor
        \Else
            \For{$j=1\dots\lfloor \frac{n}{2} \rfloor -1$}
            \Comment{Can be computed in parallel}
                \State
                $
                    \vq^{(i)}\vert_{\delta(V_{2j+1})} 
                    \gets\argmin\limits_{
                        \vmu \in \R^W
                        ~\text{s.t.}~
                        \vmu\vert_{\delta(V_{2j+1})}
                        \in \mathcal{C}_{2p+1}
                    } 
                    \KL\left(
                        \vmu\vert_{\delta(V_{2j+1})},
                        \vq^{(i-1)}\vert_{\delta(V_{2j+1})}
                    \right)$
            \EndFor
        \EndIf
    \EndFor
    \State\Return{$\vq^{(k)}$}
    \Comment{Returns an approximation of $\nabla B_Y(\vw)$}
\EndFunction
\end{algorithmic}
\caption{The \BCRF{} inference algorithm based on iterative Bregman projections, where $\vw \in \R^W$ are canonical parameters, $n$ the input length and $k$ the number of iterations. When using the $\vert$ notation on the left-hand side of an assignation, we assume only the designated coordinates are updated.}
\label{algorithm}
\end{algorithm*}

As explained in the previous section,
both MAP and marginal inference can be reduced to computing $\nabla B_Y(\cdot)$.
Next, we introduce our efficient method based on iterative Bregman projections (IBP).

First, notice that our objective can be reduced to a Kullback-Leibler (KL) projection:
\begin{align*}
    &\argmax_{\vq \in \conv(Y)}~\langle \vq, \vw \rangle + \tau H[\vq]
    \\
    &=\hspace{-5pt}\argmin_{\vq \in \conv(Y)}\langle \vq, \log \vq \rangle\!-\! \langle \vq, \vone\rangle\!-\!\langle \vq, \log \exp \tau^{-1}\vw \rangle 
    \\
    &=\hspace{-5pt}\argmin_{\vq \in \conv(Y)}~\KL[\vq | \exp \tau^{-1}\vw],
\end{align*}
where the equality in the second line holds because $\langle \vq, \vone\rangle = n\!-\!1$ is constant by constraints~\eqref{cst:enteringCluster} and~\eqref{cst:flow}.
This optimization problem aims to compute the projection of $\exp \tau^{-1}\vw$ into the set $\conv(Y)$ where the distance is measured using the KL divergence.
In order to use on IBP, we need to rewrite $\conv(Y)$ as an intersection of convex sets for which we can derive
efficient KL projections.
We describe \(\conv(Y)\) as the intersection of two polytopes \(\mathcal{C}_{\text{even}}\)
and \(\mathcal{C}_{\text{odd}}\) by partitioning the set of constraints \eqref{cst:enteringCluster} and \eqref{cst:flow} into those associated with clusters \(V_i\) with \(i\) odd, and those associated with clusters \(V_i\) with \(i\) even.\footnote{Note that clusters \(V_1\) and \(V_n\) are not considered for defining \(\mathcal{C}_{\text{even}}\) and \(\mathcal{C}_{\text{odd}}\) since all constraints are associated with a cluster \(V_i\) with \(i \in \llbracket 2, n-1 \rrbracket\).}

Let $\mathcal C_i$ be the set of vectors $\vq$ for which constraints on the arcs that are incident to vertices in cluster $V_i$ are satisfied, that is:
\begin{align*}
    \mathcal C_i = 
    \left\{
        \vq \in \R_+^E
        ~\middle|~
        \begin{array}{l}
            \sum_{a \in \delta^-(V_i)} \evq_a = 1,\\
            \forall \rv \in V_i:
            \begin{array}[t]{l}
                \sum_{a \in \delta^+(\rv)} \evq_a   \\
                 = \sum_{a \in \delta^-(\rv)} \evq_a
            \end{array}
        \end{array}
    \hspace{-0.5em}\right\}.
\end{align*}
We define \(\mathcal{C}_{\text{even}}\) and \(\mathcal{C}_{\text{odd}}\) as the intersection of polytopes \(\mathcal{C}_i\) as follows:
\begin{align*}
    \mathcal C_{\text{even}} = \bigcap_{j = 1}^{\lceil \frac{n}{2} \rceil - 1 } \mathcal C_{2j}
    \text{~~and~~}
    \mathcal C_{\text{odd}} = \bigcap_{j = 1}^{\lfloor \frac{n}{2} \rfloor -1} \mathcal C_{2j+1}.
\end{align*}
Therefore, we have $\conv(Y) = \mathcal C_{\text{even}} \cap \mathcal C_{\text{odd}}$.

We now describe how to compute the KL projection onto \( \mathcal{C}_{\text{even}}\). Since the constraints of each \( \mathcal{C}_i\) contains only variables associated with arcs \(\delta(V_i)\) and the objective function of the projection decomposes per arc, one can decompose computing \(\min_{\vq \in \mathcal C_{\text{even}}} \KL(\vq, \vw)\) as:
\[
\sum_{j=1}^{\lceil \frac{n}{2} \rceil -1} \min_{\vq\vert_{\delta(V_{2j})} \in \mathcal{C}_{2j}} \KL\left(\vq\vert_{\delta(V_{2j})},  \vw\vert_{\delta(V_{2j})} \right),
\]
where for a vector \(\vv \in \mathbb{R}^S\) and \(S' \subseteq S\), \(\vv\vert_{S'}\) denotes the the restriction of \(\vv\) to the indices associated with the elements of \(S'\), and when used in set inclusions it also restricts the right-hand side elements.
Similarly, \(\min_{\vq \in \mathcal C_{\text{odd}}} \KL(\vq, \vw)\) can be rewritten as:
\[
\sum_{j=1}^{\lfloor \frac{n}{2} \rfloor -1} \min_{\vq\vert_{\delta(V_{2j+1})} \in \mathcal{C}_{2j+1}} \hspace{-0.5em}\KL\left(\vq\vert_{\delta(V_{2j+1})},  \vw\vert_{\delta(V_{2j+1})} \right).
\]
Hence, each KL projection reduces to a sum of KL projections
\(\min_{\vq\vert_{\delta(V_i)} \in \mathcal{C}_{i}} \KL\left(\vq\vert_{\delta(V_i)},  \vw\vert_{\delta(V_i)} \right)\) 
with closed-form solutions, see
Appendix~\ref{app:IBP}.

The full algorithm simply alternates between projections in $\mathcal{C}_{\text{odd}}$ and $\mathcal{C}_{\text{even}}$.
Pseudocode is given in Algorithm~\ref{algorithm}.
A more detailed description of ITB algorithm can be found in \cite[][Sec.\ 2.1]{benamou-2015-iterat-bregm}

%% file: 4.training.tex
\section{Loss Functions}

In Section~\ref{sec:bregman},
we defined a mean regularized distribution over sequences,
and show that both marginal inference and MAP in this setting reduce to computing $\nabla B_Y(\cdot)$.
We now define loss functions for supervised and weakly-supervised learning also based on computing $\nabla B_Y(\cdot)$, and thus enjoying the same computational properties.

\subsection{Supervised Learning}

For supervised learning, we rely on the Fenchel-Young (FY) loss framework \cite{blondel2020fy}.

\begin{definition}
Given a regularization function $\Omega$, the FY loss is defined as:
\begin{align*}
\ell_{\Omega}(\vw ; \vy)
\triangleq&
- \langle \vw, \vy \rangle
+ \Omega(\vy)
\\
&+ \left(\Omega\!+\!\delta_{\conv(Y)}\right)^{*}(\vw),
\end{align*}
where $\vy \in Y$ is the gold output and $\vw$ are weights.
\end{definition}
Setting $\Omega = R$ as defined in Eq.\ \ref{eq:structured_reg},
we obtain the NLL loss for \CRF{}s since the last term is then $A_Y(\vw)$.
This framework naturally suggests using mean regularization \(\Omega = -H\) to define a loss function,
which gives the following loss and gradient:
\begin{align*}
\ell_{-H}(\vw ; \vy)
&= -\langle \vw, \vy \rangle-H(\vy)+B_Y(\vw)
\\
\text{and}~\nabla \ell_{-H}(\vw ; \vy)
&= - \vy\! + \nabla B_Y(\vw),
\end{align*}
\ie{}\ computing the gradient can be done using our inference algorithm.

\subsection{Learning with Partial Labels}

Learning with partial labels is a weakly supervised learning scenario where we do not have access to the gold annotation,
but instead, for each input, we have access to a subset of labels $\widetilde{Y} \subseteq Y$ containing the gold label.

\begin{definition}
Given a regularization function $\Omega$, the partial FY loss is defined as:
\begin{align*}
\widetilde{\ell}_{\Omega}(\vw ; \widetilde Y)
\triangleq&
\left(\Omega + \delta_{\conv(Y)}\right)^{*}(\vw)
\\
&- \left(\Omega + \delta_{\conv(\widetilde Y)}\right)^{*}(\vw),
\end{align*}
where
$\widetilde{Y} \subset Y$ is a non-empty set containing the expected gold output.
\end{definition}
\begin{thm}\label{prop:partial_fy}
Partial FY losses satisfy the following properties:
\begin{enumerate}[leftmargin=*]
    \item \textit{Generalization of FY losses.}
        If $\widetilde Y\!=\!\{ \vy \}$ is a singleton containing only a gold label, then
        \(
        \widetilde{\ell}_{\Omega}(\vw, \{ \vy \}) = \ell_\Omega(\vw, \vy).
        \)
    \item \textit{Non-negativity.}
        The loss is bounded below by 0.

    \item \textit{Smaller partial labeling set $\implies$ bigger loss.}
        Let $\widetilde Y' \subseteq \widetilde Y$, then:
        \[
            \widetilde{\ell}(\vw ; \widetilde Y') \geq \widetilde{\ell}(\vw ; \widetilde Y).
        \]
\end{enumerate}
\end{thm}
Proofs are given in App.~\ref{app:loss}.
This family of losses was introduced by \citet{stewart2013diffclustering}.
They analyze properties of the loss function when $\Omega = 0$, that we (trivially) extend to more general regularization.
Importantly, property (5) shows that the more information we have, the better is the loss

Setting $\Omega = R$, we obtain the NLL loss after marginalizing over $\widetilde{Y}$, a standard approach in learning from partial labels \cite{jin2002partial} that can be understood as a special instance of the EM algorithm, see \cite[Sec.\ 4.1]{corro2024discner}.
\citet{yang2018partialner} and \citet{huang-etal-2019-learning} learned sequence labeling models using this loss.
For the \BCRF{} case, we rely on mean regularization and define the following loss function and gradient:
\begin{align*}
\widetilde{\ell}_{-H}(\vw ; \widetilde{Y})\!
&= B_Y(\vw)\!-\! B_{\widetilde{Y}}(\vw)
\\
\text{and}~\nabla\widetilde{\ell}_{-H}(\vw ; \widetilde{Y})\!
&= \nabla B_Y(\vw)\!-\! \nabla B_{\widetilde{Y}}(\vw)
\end{align*}
which requires two calls to our IBP algorithm.

%% file: 5.related_work.tex
\section{Related Work}

The main motivation of our approach is to develop a sequence labeling model suitable for parallel processors, e.g.\ GPUs.
Developing such models is a shared preoccupation in deep learning, starting with the ubiquitous transformers~\cite{nips2017_3f5ee243} or more recently state-space models~\cite{gu-2022-effic-model} and following works on linear transformers \cite[][Table~2]{yang2024deltarule}.
Designing NLP algorithms for GPUs has been explored in syntactic parsing~\cite{johnson-2011-parsing,hall-etal-2014-sparser} and machine translation~\cite{he-etal-2015-gappy,argueta-chiang-2018-composing}.
More generally, efficient parallelization of dynamic algorithms is widely studied~\cite{valiant-1983-fast-paral,maleki-2014-paral}.

With motivations similar to ours,
\citet{wang-etal-2020-ain}
proposed to rely on mean field \cite[\MF{},][]{weiss1907mft,parisi1979mft} to approximate a \CRF{} distribution using a parallel algorithm.
However, \MF{} has the following issues, all of which are solved by our approach:
(1)~the \MF{} objective (see Appendix~\ref{sec:mean-field-theory} for details) is non-convex \cite[Sec.~5.4]{wainwright2008expfam}, and therefore its approximation strongly depends on initialization;
(2)~the parallel decoding algorithm used for \MF{} by \citet{wang-etal-2020-ain} is not even guaranteed to converge to a local minima \cite{kraehenbuehl13mftconvergence};
(3)~\MF{} cannot handle highly structured problems, \ie{}\ problems where there are forbidden transitions between tags to ensure well-formedness of prediction (see our experiments and Appendix~\ref{sec:mean-field-theory}).

Our method also has connections with frameworks based on mean regularization, such as \textsc{SparseMAP}~\cite{niculae-2018-spars}.
The main difference is that we use an entropic regularization in order to develop a fast alternative to the dynamic programming (DP) inference algorithms,
while \textsc{SparseMAP} solver relies on iterative calls to these DP algorithms, and hence is slower.

%% file: 6.experiments.tex
\begin{table*}
    \null\hfill%
    \begin{subtable}[t]{0.65\textwidth}
        \setlength{\tabcolsep}{2pt}
        \centering
        \footnotesize
        \input{results/pos_accuracy_1}
        \caption{Accuracy results for POS tagging.}
        \label{tab:pos}
    \end{subtable}%
    \hfill%
    \begin{subtable}[t]{0.22\textwidth}
        \setlength{\tabcolsep}{2pt}
        \centering
        \footnotesize
        \input{results/ner_supervised_eng_1}
        \caption{NER F-scores.}
        \label{tab:ner_super}
    \end{subtable}%
    \hfill\null
    \vspace{0.8em}
    \null\hfill%
    \begin{subtable}[t]{0.65\textwidth}
        \setlength{\tabcolsep}{2pt}
        \centering
        \footnotesize
        \input{results/joint_seg_tagging_f1_1}
        \caption{F-score for joint segmentation and POS tagging.}
        \label{tab:seg}
    \end{subtable}%
    \hfill%
    \begin{subtable}[t]{0.22\textwidth}
        \setlength{\tabcolsep}{2pt}
        \centering
        \footnotesize
        \input{results/ner_partial_1}
        \caption{Partially supervised NER F-scores.}
        \label{tab:ner_partial}
    \end{subtable}%
    \hfill\null
    \vspace{-0.5em}
    \caption{Experimental results. The unstructured model does not have transition scores neither constraints between adjacent pair of tags. For each training approach, we evaluate different decoding strategies.}
\end{table*}

\begin{table*}
    \null\hfill%
    \begin{subtable}[t]{0.45\textwidth}
        \setlength{\tabcolsep}{2pt}
        \centering
        \footnotesize
        \input{timing/pos_training_simple}
        \caption{POS tagging.}
        \label{tab:pos_speed}
    \end{subtable}%
    \hfill%
    \begin{subtable}[t]{0.25\textwidth}
        \setlength{\tabcolsep}{2pt}
        \centering
        \footnotesize
        \input{timing/joint_training}
        \caption{Joint word segmentation and POS tagging.}
        \label{tab:tok_speed}
    \end{subtable}%
    \hfill%
    \begin{subtable}[t]{0.18\textwidth}
        \setlength{\tabcolsep}{2pt}
        \centering
        \footnotesize
        \input{timing/ner_training}
        \caption{NER.}
        \label{tab:ner_speed}
    \end{subtable}%
    \hfill\null
    \vspace{-0.5em}
    \caption{Relative training time speed-up compared to using the forward algorithm to compute the loss. We set the number of iterations to 10. Column \emph{L.}\ is the number of encoder layers, whereas \emph{Base}/\emph{Large} refer to \textsc{BERT} sizes.}
    \label{tab:speed}
\end{table*}

\section{Experiments}%
\label{sec:experiments}

We present a series of experiments on part-of-speech (POS) tagging, joint word segmentation and POS tagging, and named entity recognition (NER). We compare \CRF{} (\ie{} Viterbi/forward), \MF{} and \BCRF{} in terms of accuracy and time.
For reference, we also include an unstructured model, that predict each tag independently.\footnote{Unstructured models cannot guarantee well-formedness of predictions for highly constrained settings. In these cases, we rely on simple heuristics to construct valid outputs.}
For \MF{} and \BCRF{}, we evaluate the algorithms with both 5 and 10 iterations of the respective inference algorithms.
For decoding with \BCRF{}, we set $\tau^{-1} = 10$ in all experiments.

We use datasets from the Universal Dependency Treebank~2.15~\cite[UD,][]{nivre-etal-2020-universal} for the first two tasks and the \textsc{CoNLL} 2003 shared task data~\cite{tjong-kim-sang-de-meulder-2003-introduction} for the last one.
We use standard splits in all cases.
We average results over 5 runs initialized randomly.

\paragraph{Neural network.}
The weight function $f_{\theta}$ is implemented by a self-attentive encoder network \cite{nips2017_3f5ee243} for all our experiments.
Tag and transition scores are computed by a simple multi-linear perceptron.
Note that our model is non-homogeneous,
that is transition scores are computed independently for each adjacent word.
The neural network is trained from scratch, except for NER where we use \textsc{Bert}~\cite{devlin-etal-2019-bert}.
Hyperparameters are given in Appendix~\ref{sec:train-impl}.

\subsection{Part-of-Speech Tagging}%
\label{sec:exp_pos}

We evaluate on four corpora from UD: LassySmall (Dutch), EWT (English) and GSD (French and German).
We use coarse POS tags for all languages except French as it contains only UD tags.
Results are given in Table~\ref{tab:pos}.
We can see that all methods are close, but our algorithm outperforms \MF{} for approximate decoding after training with the standard \CRF{} loss.

\subsection{Word Segmentation and POS Tagging}%
\label{sec:exp_segmentation}

We cast joint word segmentation and POS tagging as a character sequence labeling problem in the BIES format \cite{ratinov2009ner}.
For example, tag \texttt{B-VB} indicates the first character of a multi-character verb, whereas \texttt{S-VB} indicates a single character verb.
Note that this problem is highly structured: for example, we can only assign tag \texttt{I-VB} if the previous tag was \texttt{B-VB} or \texttt{I-VB}.
This is implemented using $-\infty$ weights on forbidden tag transitions.
We report the F-score on reconstructed tagged words.
We present experiments on Chinese and Japanese UD GSD datasets.

Results are given in Table~\ref{tab:seg}.
We see that on this task, with more tags than simple POS tagging and, more importantly, with constraints on forbidden adjacent tag pairs, both unstructured approach and \MF{} struggle to recover the exact \CRF{} performance.
In contrast, decoding with the \BCRF{} approach performs better than \MF{}, and when training is performed with \BCRF{} too, the F-scores are almost similar to the \CRF{} model.

\subsection{Named-Entity Recognition}%
\label{sec:exp_ner}

For NER we report F-scores of reconstructed labeled mentions on \textsc{CoNLL} 2003, using \textsc{Bert}~\cite{devlin-etal-2019-bert} in the base and large versions.
In this series of experiments we test supervised and partially supervised training regimes.

For supervised training, we report results in Table~\ref{tab:ner_super}.
As in POS tagging, we see that \BCRF{} is closer to \CRF{} than \MF{}, and in the case of the large \textsc{Bert} model, \BCRF{} manages to (slightly) improve over \CRF{} decoding.

For partially supervised learning, we follow the same setting as \citet{huang-etal-2019-learning} and split the training set into 4 subsets, each annotated with only one entity type.
For a given sentence $\vs$, the partial labeling $\widetilde{Y}$ is all sequence labelings that contains at least the gold annotated tags.
We report results in Table~\ref{tab:ner_partial}.
We see that our method recovers the same accuracy as \CRF{} when trained with \CRF{} loss, while \MF{} does not.
When trained with \BCRF{}, there is a small drop in performance, but \CRF{} and \BCRF{} give equivalent results while \MF{} is lower.

\subsection{Speed Improvement}
\label{sec:time-speed-up}

\begin{table}
    \null\hfill%
    \begin{subtable}[t]{0.94\columnwidth}
        \setlength{\tabcolsep}{2pt}
        \centering
        \footnotesize
        \input{timing/pos_decoder_timing}
        \caption{POS tagging}
        \label{table:decoder_timing_pos}
    \end{subtable}%
    \hfill\null\vspace{0.8em}
    
    \null\hfill%
    \begin{subtable}[t]{0.55\columnwidth}
        \setlength{\tabcolsep}{2pt}
        \centering
        \footnotesize
        \input{timing/joint_decoder_timing}
        \caption{Joint word segmentation and POS tagging.}
        \label{table:decoder_timing_joint}
    \end{subtable}%
    \hfill%
    \begin{subtable}[t]{0.34\columnwidth}
        \setlength{\tabcolsep}{2pt}
        \centering
        \footnotesize
        \input{timing/ner_decoder_timing}
        \caption{NER.}
        \label{table:decoder_timing_ner}
    \end{subtable}%
    \hfill\null
    \vspace{-0.5em}
    \caption{Relative decoding time speed-up compared to the Viterbi algorithm. Column \emph{L.}\ is the number of encoder layers, whereas \emph{Base}/\emph{Large} refer to \textsc{BERT} sizes.}
    \label{tab:decoder_timing}
\end{table}

We now analyze the time speed-up.
All timing include the forward (and backward if training) pass in the neural network.

We report relative training time in Table~\ref{tab:speed}.
For POS tagging our model trains up to 7.3 times faster than standard \CRF{} training with simple word embeddings and up to 4.7 times when using 2 encoder layers.
Moreover, our method is faster than \MF{} training. 
For joint segmentation and POS tagging, \MF{} is slightly faster, but at the expense of a drop in performance (see Table~\ref{tab:seg}).
Importantly, for NER, \MF{} cannot be used in the partial supervision scenario.
Our \BCRF{} approach is significantly faster than \CRF{} training even when using a full \textsc{Bert} model.

We report relative decoding time in Table~\ref{tab:decoder_timing}.
Although \MF{} is faster that \BCRF{}, the difference diminishes as the network becomes larger.
Moreover, as shown in downstream task results, \BCRF{} can handle structual constraint, while still being faster than using the Viterbi algorithm for decoding.

%% file: results/pos_accuracy_1.tex
\begin{tabular}{llcccccccc}
\toprule
&& \multicolumn{4}{c}{No self-attentive encoder} & \multicolumn{4}{c}{2 layers}\\
\cmidrule(lr){3-6}\cmidrule(lr){7-10}
& & Dutch & English & French & German & Dutch & English & French & German\\
\midrule
\multicolumn{10}{l}{\textbf{Baseline}} \\
\midrule
\multicolumn{2}{l}{Unstructured} & 90.5 & 85.8 & 92.7 & 89.0 & 93.4 & 90.8 & 96.0 & 94.0\\
\midrule
\multicolumn{10}{l}{\textbf{\CRF{} training}} \\
\midrule
\multicolumn{2}{l}{Viterbi}   & 94.3 & 91.3 & 95.9 & 94.1 & 94.7 & 91.9 & 96.2 & 94.3\\
\addlinespace[0.3em]
\MF & 5 it.  & 93.8 & 90.6 & 95.5 & 94.0 & 94.4 & 91.0 & 95.8 & 94.2\\
\MF & 10 it.  & 93.9 & 90.7 & 95.5 & 94.1 & 94.5 & 91.0 & 95.8 & 94.2\\
\addlinespace[0.3em]
\BCRF{} & 5 it.  & 94.4 & 91.3 & 95.9 & 94.1 & 94.7 & 91.9 & 96.2 & 94.3\\
\BCRF{} & 10 it.  & 94.3 & 91.3 & 95.9 & 94.1 & 94.7 & 91.9 & 96.2 & 94.3\\
\midrule
\multicolumn{10}{l}{\textbf{\MF{} training}} \\
\midrule
\multicolumn{2}{l}{Viterbi}   & 93.7 & 89.1 & 95.8 & 93.8 & 94.6 & 91.6 & 96.3 & 94.0\\
\addlinespace[0.3em]
\MF & 5 it.   & 94.3 & 91.4 & 95.9 & 94.0 & 94.7 & 91.7 & 96.3 & 94.1\\
\MF & 10 it.  & 94.3 & 91.4 & 95.9 & 94.0 & 94.7 & 91.7 & 96.3 & 94.1\\
\midrule
\multicolumn{10}{l}{\textbf{\BCRF{} training}} \\
\midrule
\multicolumn{2}{l}{Viterbi}   & 94.1 & 91.1 & 95.8 & 94.0 & 94.6 & 91.8 & 96.5 & 94.4\\
\addlinespace[0.3em]
\BCRF{} & 5 it.  & 94.1 & 91.1 & 95.8 & 94.0 & 94.6 & 91.8 & 96.5 & 94.4\\
\BCRF{} & 10 it.  & 94.1 & 91.1 & 95.8 & 94.0 & 94.6 & 91.8 & 96.5 & 94.4\\
\bottomrule
\end{tabular}

%% file: results/ner_supervised_eng_1.tex
\begin{tabular}{llcc}
\toprule
& & \multicolumn{2}{c}{\textsc{Bert}}\\
\cmidrule(lr){3-4}
& & Base & Large\\
\midrule
\multicolumn{4}{l}{\textbf{Baseline}} \\
\midrule
\multicolumn{2}{l}{Unstructured}    & 91.6 & 92.0\\
\midrule
\multicolumn{4}{l}{\textbf{\CRF{} training}} \\
\midrule
\multicolumn{2}{l}{Viterbi}  & 91.9 & 92.3\\
\addlinespace[0.3em]
\MF{} & 5 it.  & 91.7 & 91.9\\
\MF{} & 10 it.  & 91.6 & 91.9\\
\addlinespace[0.3em]
\BCRF{} & 5 it.  & 91.9 & 92.3\\
\BCRF{} & 10 it.  & 91.9 & 92.3\\
\midrule
\multicolumn{4}{l}{\textbf{\MF{} training}} \\
\midrule
\multicolumn{2}{l}{Viterbi} & 91.7 & 92.2\\
\addlinespace[0.3em]
\MF{} & 5 it.   & 91.6 & 92.1\\
\MF{} & 10 it.  & 91.6 & 92.1\\
\midrule
\multicolumn{4}{l}{\textbf{\BCRF{} training}} \\
\midrule
\multicolumn{2}{l}{Viterbi}  & 91.7 & 92.1\\
\addlinespace[0.3em]
\BCRF{} & 5  & 91.6 & 92.1\\
\BCRF{} & 10  & 91.7 & 92.1\\
\bottomrule
\end{tabular}

%% file: results/joint_seg_tagging_f1_1.tex
\begin{tabular}{lrcccccc}
\toprule
&& \multicolumn{2}{c}{No self-att.\ encoder} & \multicolumn{2}{c}{2 layers} & \multicolumn{2}{c}{4 layers}\\
\cmidrule(lr){3-4}\cmidrule(lr){5-6}\cmidrule(lr){7-8}
& & Chinese & Japanese & Chinese & Japanese & Chinese & Japanese\\
\midrule
\multicolumn{8}{l}{\textbf{Baseline}} \\
\midrule
\multicolumn{2}{l}{Unstructured}  & 47.3 & 52.9 & 62.0 & 86.3 & 64.9 & 89.0\\
\midrule
\multicolumn{8}{l}{\textbf{\CRF{} training}} \\
\midrule
\multicolumn{2}{l}{Viterbi}   & 84.2 & 90.6 & 84.3 & 90.8 & 84.3 & 90.7\\
\addlinespace[0.3em]
\MF{} & 5  & 72.9 & 76.5 & 71.4 & 75.4 & 71.9 & 75.1\\
\MF{} & 10  & 74.6 & 77.6 & 72.2 & 76.1 & 72.5 & 76.0\\
\addlinespace[0.3em]
\BCRF{} & 5  & 81.9 & 88.7 & 81.7 & 88.7 & 82.0 & 88.5\\
\BCRF{} & 10  & 83.7 & 90.1 & 83.7 & 90.4 & 83.8 & 90.2\\
\midrule
\multicolumn{8}{l}{\textbf{\MF{} training}} \\
\midrule
\multicolumn{2}{l}{Viterbi}    & 77.6 & 79.6 & 79.9 & 82.3 & 81.1 & 83.5\\
\addlinespace[0.3em]
\MF{} & 5  & 81.0 & 88.7 & 80.7 & 88.8 & 80.8 & 89.0\\
\MF{} & 10  & 82.0 & 89.1 & 81.2 & 89.1 & 81.2 & 89.3\\
\midrule
\multicolumn{8}{l}{\textbf{\BCRF{} training}} \\
\midrule
\multicolumn{2}{l}{Viterbi}    & 83.6 & 89.8 & 84.5 & 90.5 & 84.0 & 90.7\\
\addlinespace[0.3em]
\BCRF{} & 5  & 82.2 & 88.8 & 83.3 & 89.8 & 82.8 & 90.1\\
\BCRF{} & 10  & 83.2 & 89.6 & 84.2 & 90.4 & 83.8 & 90.6\\
\bottomrule
\end{tabular}


%% file: results/ner_partial_1.tex
\begin{tabular}{llcc}
\toprule
& & \multicolumn{2}{c}{\textsc{Bert}}\\
\cmidrule(lr){3-4}
& & Base & Large\\
\midrule
\multicolumn{4}{l}{\textbf{\CRF{} training}} \\
\midrule
\multicolumn{2}{l}{Viterbi}  & 90.5 & 91.2\\
\addlinespace[0.3em]
\MF{} & 5 it. & 90.2 & 90.8\\
\MF{} & 10 it. & 90.3 & 90.8\\
\addlinespace[0.3em]
\BCRF{} & 5 it. & 90.4 & 91.2\\
\BCRF{} & 10 it. & 90.5 & 91.2\\
\midrule
\multicolumn{4}{l}{\textbf{\BCRF{} training}} \\
\midrule
\multicolumn{2}{l}{Viterbi}  & 90.6 & 91.2\\
\addlinespace[0.3em]
\MF{} & 5 it. & 90.3 & 90.9\\
\MF{} & 10 it. & 90.3 & 90.9\\
\addlinespace[0.3em]
\BCRF{} & 5 it. & 90.6 & 91.2\\
\BCRF{} & 10 it.  & 90.6 & 91.2\\
\bottomrule
\end{tabular}

%% file: timing/pos_training_simple.tex
\begin{tabular}{rccccccccc}
\toprule
 & \multicolumn{2}{c}{Dutch} & \multicolumn{2}{c}{English} & \multicolumn{2}{c}{French} & \multicolumn{2}{c}{German}\\
 \cmidrule(lr){2-3}\cmidrule(lr){4-5}\cmidrule(lr){6-7}\cmidrule(lr){8-9}
L. &\textsc{Mf} & \textsc{Bcrf} & \textsc{Mf} & \textsc{Bcrf} & \textsc{Mf} & \textsc{Bcrf} & \textsc{Mf} & \textsc{Bcrf} \\
\midrule
0 & $\times6.2$ & $\times7.3$ & $\times5.0$ & $\times6.1$ & $\times6.0$ & $\times6.5$ & $\times5.4$ & $\times6.6$ \\
2 & $\times3.8$ & $\times4.1$ & $\times3.9$ & $\times4.1$ & $\times4.3$ & $\times4.7$ & $\times3.8$ & $\times4.0$ \\
\bottomrule
\end{tabular}

%% file: timing/joint_training.tex
\begin{tabular}{lcccc}
\toprule
& \multicolumn{2}{c}{Chinese} & \multicolumn{2}{c}{Japanese} \\
\cmidrule(lr){2-3}\cmidrule(lr){4-5}
L. & \MF & \BCRF & \MF & \BCRF \\
\midrule
0 & $\times8.8$ & $\times7.8$ & $\times12.1$ & $\times12.3$ \\
2 & $\times5.5$ & $\times5.4$ & $\times5.6$  & $\times5.9 $\\
4 & $\times4.1$ & $\times4.0$ & $\times4.4$  & $\times4.5 $\\
\bottomrule
\end{tabular}

%% file: timing/ner_training.tex
\begin{tabular}{lcc}
\toprule
& Sup. & Partial \\
\cmidrule(lr){2-2}\cmidrule(lr){3-3}
& \BCRF{} & \BCRF{} \\
\midrule
Base  & $\times 1.9$ & $\times 1.7$ \\
Large & $\times 1.2$ & $\times 1.2$ \\
\bottomrule
\end{tabular}

%% file: timing/pos_decoder_timing.tex
\begin{tabular}{rccccccccc}
\toprule
 & \multicolumn{2}{c}{Dutch} & \multicolumn{2}{c}{English} & \multicolumn{2}{c}{French} & \multicolumn{2}{c}{German}\\
 \cmidrule(lr){2-3}\cmidrule(lr){4-5}\cmidrule(lr){6-7}\cmidrule(lr){8-9}
L. &\textsc{Mf} & \textsc{Bcrf} & \textsc{Mf} & \textsc{Bcrf} & \textsc{Mf} & \textsc{Bcrf} & \textsc{Mf} & \textsc{Bcrf} \\
\midrule
0 & $\times9.1$ & $\times5.4$ & $\times8.0$ & $\times4.9$ & $\times10.4$ & $\times7.0$ & $\times8.5$ & $\times5.2$ \\
2 & $\times6.6$ & $\times4.7$ & $\times6.2$ & $\times4.4$ & $\times8.1$ & $\times5.9$ & $\times6.4$ & $\times4.5$ \\
\bottomrule
\end{tabular}

%% file: timing/joint_decoder_timing.tex
\begin{tabular}{lcccc}
\toprule
& \multicolumn{2}{c}{Chinese} & \multicolumn{2}{c}{Japanese}
\\
\cmidrule(lr){2-3}\cmidrule(lr){4-5}
L. & \MF{} & \BCRF{} & \MF{} & \BCRF{}
\\
\midrule
\multicolumn{5}{l}{\textbf{5 iterations}}
\\
\midrule
0
& $\times 14.2$ & $\times 7.7$
& $\times 24.6$ & $\times 14.6$
\\
2
& $\times 9.0$ & $\times 6.2$
& $\times 12.0$ & $\times 9.3$
\\
4
& $\times 6.9$ & $\times 5.0$
& $\times 9.3$ & $\times 7.8$
\\
\midrule
\multicolumn{5}{l}{\textbf{10 iterations}}
\\
\midrule
0
& $\times 9.5$ & $\times 5.2$
& $\times 19.8$ & $\times 11.0$
\\
2
& $\times 7.1$ & $\times 4.6$
& $\times 10.5$ & $\times 7.7$
\\
4
& $\times 5.8$ & $\times 3.8$
& $\times 8.7$ & $\times 6.8$
\\
\bottomrule
\end{tabular}

%% file: timing/ner_decoder_timing.tex
\begin{tabular}{lcc}
\toprule
& \MF{} & \BCRF{}
\\
\midrule
\multicolumn{3}{l}{\textbf{5 iterations}}
\\
\midrule
Base
& $\times 2.9$ & $\times 2.6$
\\
Large
& $\times 1.6$ & $\times 1.5$
\\
\midrule
\multicolumn{3}{l}{\textbf{10 iterations}}
\\
\midrule
Base
& $\times 2.9$ & $\times 2.4$
\\
Large
& $\times 1.6$ & $\times 1.5$
\\
\bottomrule
\end{tabular}

%% file: 7.conclusion.tex
\section{Conclusion}

In this work, we introduce \BCRF{}, a novel sequence labeling model based on entropic mean regularization, and a novel inference algorithm based on iterative Bregman projections.
This method is designed to take full advantage of parallel processors.

While we designed \BCRF{} for sequence labeling, we believe that the proposed methodology paves the way for novel inference algorithms in other structured prediction settings, such as parsing but also automatic speech recognition for which Viterbi and forward are known bottlenecks \cite{ondel2022mmi}.

%% file: 8.limitations.tex
\section*{Limitations}

While our model demonstrates that a theoretical treatment of a NLP task deemed simple and well understood can lead to a novel method with practical effectiveness, it suffers some limitations.

First, our model is designed with parallel computing architectures in mind, \ie{}\ GPUs. On a purely sequential architecture.
However, we observe that parallel architectures have become ubiquitous and so it seems a reasonable limitation.

Second, our method relies on the fact that the continuous relaxation of the optimization problem underlying the labeling task is integral.
While our approach can be easily adapted to second-order linear chains using the standard trick \cite{he1998secondorder},
it cannot directly rely on the \emph{factorized} second-order \CRF{} trick of~\citet{wang-etal-2020-ain}.
More generally, our approach cannot be applied on \CRF{} more general than linear chain ones as our formulation would not ensure local consistency without extra variables, but adding these variables results in problems for which we may not find a closed-form expression of the solution to the KL projection. However, our method could be used to solve linear-chain \CRF{} arising from dual decomposition based methods like in the row and column decomposition of a grid \CRF{} \citep{komodakisMRFEnergy}.

Finally, while our approach is generic in the sense that it could be applied to any dynamic programming algorithm that operates on graphs, it cannot be trivially adapted to dynamic programming algorithms that operate on hypergraphs \cite{martin1990polyhedral}.

%% file: 9.dp.tex
\section{Dynamic Programming Recursion}
\label{app:dp}

This section contains proof of correctness for both Viterbi and forward algorithms in a single analysis of the dynamic programming recursion.
A popular technique to this end is based on the semiring parsing framework \cite{goodman1999semiring}.
We give an alternate analysis by following the analysis of \citet{mensch2018diffdp}.\footnote{More precisely, what we describe here is exactly their proof, but under our notation to make our paper self-contained for newcomers.}

We first prove two properties of the $\maxOmega$ operator for $\Omega \in \{0, -H\}$: associativity, and distributivity over addition.

\subsection{Associativity of $\maxOmega$}

We show that:
\[
\maxOmega\begin{bmatrix}
    \maxOmega(\vv)\\
    \maxOmega(\vv')
\end{bmatrix}
=
\maxOmega(\vu)
\]
for all $\vv \in \R^d, \vv' \in \R^{d'}$ and $\vu \in \R^{d+d'}$ where:
\[
\evu_i \triangleq \begin{cases}
    \evv_i \quad&\text{if } i \in \llbracket 1, d \rrbracket\,,\\
    \evv'_{i - d}&\text{otherwise,}
\end{cases}
\]
i.e. $\vu$ is the concatenation of $\vv$ and $\vv'$.

In the case $\Omega = 0$,
we have:
\begin{align*}
    {\max\displaystyle_0}\begin{bmatrix}
        {\max\displaystyle_0} \vv, \\
        {\max\displaystyle_0} \vv'
    \end{bmatrix}
    &= {\max\displaystyle_0}
        \begin{bmatrix}
            \max_i(\evv_i) \\
            \max_i(\evv'_i)
        \end{bmatrix}
    \\
    &= \max_i(\evu_i) \\
    &= {\max\displaystyle_0}(\vu)
\end{align*}
which proves the property.

In the case $\Omega = -H$,
we have:
\begin{align*}
    &{\max\displaystyle_{-H}}\begin{bmatrix}
        {\max\displaystyle_{-H}} \vv, \\
        {\max\displaystyle_{-H}} \vv'
    \end{bmatrix}
    \\
    &= {\max\displaystyle_{-H}}
        \begin{bmatrix}
            \log\sum_i \exp(\evv_i) \\
            \log\sum_i \exp(\evv'_i)
        \end{bmatrix}
    \\
    &= \log\left(
        \begin{array}{l}\displaystyle
            \cancel{\exp\log}\sum_i \exp(\evv_i)\\
            +
            \cancel{\exp\log}\sum_i \exp(\evv'_i)
        \end{array}
        \right)
    \\
    &= \log\sum_i \exp(\evu_i) \\
    &= {\max\displaystyle_{-H}}(\vu)
\end{align*}
which proves the property.

\subsection{Distributivity over addition of $\maxOmega$}
We show that:
\[
\maxOmega(\vv + c \vone) = c + \maxOmega(\vv)\,,
\]
for all vectors $\vv \in \R^d$ and scalars $c \in \R$.

In the case $\Omega = 0$,
we  have:
\[
{\max\displaystyle_0} (\vv + c \vone)
= \max_{i} (\evv_i + c)
= c + {\max\displaystyle_0}(\vv)\,,
\]
which proves the property.

In the case $\Omega = -H$,
we have:
\begin{align*}
{\max\displaystyle_{-H}}(\vv + c \vone)
&=\log \sum_{i} \exp(\evv_i + c) \\
&=\log \sum_{i} \exp(\evv_i)\exp(c) \\
&=\log \left(\exp(c) \sum_{i} \exp(\evv_i)\right) \\
&=c + \log\sum_{i} \exp(\evv_i)\\
&=c + {\max\displaystyle_{-H}}(\vv)\,,
\end{align*}
which proves the property.

\subsection{Correctness Proof}

We now turn to the proof of the recursion.\footnote{We stress out that this proof is only true for $\Omega \in \{0, -H\}$, see \cite[Proposition~2]{mensch2018diffdp}}
By definition, we have:
\begin{align*}
    &c_{i,t}(\vw)
    = \maxOmega
        \left[
            \sum_{j=1}^{i-1} \langle \vw_j, \phi(\vx)_j \rangle
        \right]_{\vx \in X_i | \evx_i = t},
    \\
\intertext{where the regularized maximum is taken on all tag sequences from the beginning until position $i$, where position $i$ is constrained to be tagged with $t \in T$. We can split the sum by extracting the term that weights the last transition:}
    &= \maxOmega
        \left[
            \begin{array}{l}\displaystyle
            \sum_{j=1}^{i-2} \langle \vw_j, \phi(\vx)_j \rangle \\
            + \langle \vw_{i-1}, \phi(\vx)_{i-1} \rangle
            \end{array}
        \right]_{\vx \in X_i | \evx_i = t},
    \\
\intertext{which can be equivalently written as a regularized maximum over sequences in $X_{i-1}$ as follows:}
    &= \maxOmega
        \left[
            \begin{array}{l}\displaystyle
            \sum_{j=1}^{i-2} \langle \vw_j, \phi(\vx)_j \rangle \\
            + \vw_{i-1, x_{i-1}, t}
            \end{array}
        \right]_{\vx \in X_{i-1}}.
    \\
\intertext{Thank to associativity property of $\maxOmega$, we can ``split'' the operation over $|T|$ regularized maximums plus one that aggregates the results:}
    &= \maxOmega
        \left[
            \maxOmega
                \left[
                    \begin{array}{l}\displaystyle
                        \sum_{j=1}^{i-2} \langle \vw_j, \phi(\vx)_j \rangle \\
                        + \vw_{i-1, t', t}
                    \end{array}
                \right]_{\substack{\vx \in X_{i-1}\\| \evx_{i-1} = t'}}
            \right]_{t' \in T}\hspace{-0.6cm}.
    \\
\intertext{Finally, using distributivity over addition we extract the term $\vw_{i-1, t', t}$ and obtain the recursive formulation used by the dynamic programming algorithm:}
    &= \maxOmega
        \left[\hspace{-0.1cm}
            \begin{array}{l}\displaystyle
                \vw_{i-1, t', t}\\
                + \maxOmega
                    \left[\displaystyle
                        \sum_{j=1}^{i-2} \langle \vw_j, \phi(\vx)_j \rangle
                    \right]_{\substack{\vx \in X_{i-1}\\| \evx_{i-1} = t'}}
            \end{array}\hspace{-0.1cm}
        \right]_{t' \in T}
    \\
    &= \maxOmega\left[ \vw_{i-1, t', t}
        + c_{i-1, y'}(\vw) \right]_{t' \in T}.
\end{align*}

%% file: 9.logsumexp.tex
\section{Variational Formulation of the Log-Partition Function}
\label{app:logsumexp}

Let $\vv \in \R^d$ be a vector.
In this appendix, we prove the variational formulation of the log-partition function, that is:
\begin{align*}
    \log \sum_i \exp \evv_i
    &=
    \max_{\vmu \in \triangle(d)}~\langle \vv, \vmu  \rangle - \langle \vmu, \log \vmu\rangle
    \\
    &= \textstyle{\max_{-H}}(\vv),
\end{align*}
or, in other words, that the Shannon entropy function restricted to the simplex is the Fenchel conjugate of the \emph{logsumexp} function.
This results is well-known, see \cite[Example~3.25]{boyd2004convex} and \cite[][Section~4.4.10]{beck2017firstorder}, among others.

We start by explicitly writing the mathematical program:
\begin{align*}
    \max_{\vmu \in \R^d}~
    &\langle \vv, \vmu  \rangle - \langle \vmu, \log \vmu\rangle
    \\
    \text{s.t.}~
    & \langle \vmu, \vone\rangle = 1\\
    & \vmu \geq \vzero
\end{align*}
Dualizing the constraints gives the following Lagrangian function:
\begin{align*}
    \mathcal L(\vmu, \lambda, \vnu) =
    & \langle \vv, \vmu  \rangle - \langle \vmu, \log \vmu\rangle \\
    & + \lambda (1 - \langle \vmu, \vone\rangle) + \langle \vnu, \vmu\rangle,
\end{align*}
where $\lambda \in \R$ and $\vnu \in \R^d_{\geq 0}$ are dual variables associated with the equality and inequalities, respectively.

As the objective is concave and all constraints are linear, $\widehat \vmu$, $\widehat \lambda$ and $\widehat \vnu$ are optimal primal and dual variables if and only if they satisfy the KKT conditions \cite[Sec.\ 5.5.3]{boyd2004convex}.
By stationarity, we have:
\begin{align*}
    \frac{\partial}{\partial \widehat\mu_i}  \mathcal L(\widehat\vmu, \widehat\lambda, \widehat\vnu) &= 0
    \\
    \log \widehat\mu_i
    &= \evv_i - \widehat\lambda + \widehat\mu_i
    \\
    \widehat\mu_i &= \exp(\evv_i - \widehat\lambda + \widehat\mu_i).
\intertext{Note that $\widehat\mu_i > 0$ and by complementary slackness we must have $\widehat\nu_i \widehat\mu_i = 0$, therefore $\widehat\nu_i = 0$, and we can write:}
    \widehat\mu_i &= \frac{\exp(\evv_i)}{\exp(\widehat\lambda)}.
\end{align*}
By primal feasilibty, we have:
\begin{align*}
    \langle \widehat\vmu, \vone\rangle &= 1
    \\
    \sum_i \frac{\exp(\evv_i)}{\exp(\widehat\lambda)} &= 1
    \\
    \exp(\widehat\lambda) = \sum_i \exp(\evv_i),
\end{align*}
and therefore:
\begin{align*}
    \widehat\mu_i = \frac{\exp(\evv_i)}{\sum_j \exp(\evv_j)}.
\end{align*}
Plugging the optimal primal variables in the objective, we obtain:
\begin{align*}
    \langle \vv, \widehat\vmu  \rangle - \langle \widehat\vmu, \log \widehat\vmu\rangle
    &= \log \sum_i \exp \evv_i.
\end{align*}

%% file: 9.subgradient_fenchel_conjugate.tex
\section{Subgradient of the Fenchel Conjugate}
\label{app:subgradient}

First, note that Eq.\ \eqref{eq:structured_reg_grad} can be rewritten as the gradient of a Fenchel conjugate.
Indeed, the log-partition function can be rewritten as:
\begin{align*}
    A_Y(\vw)
    &= \max_{\vq \in \conv(Y)}~\langle \vq, \vw \rangle - R(\vq)
    \\
    &=
    \left(-R + \delta_{\conv(Y)}\right)^*(\vw).
\end{align*}
Therefore, we give a simple proof of the (sub)gradient of the Fenchel conjugate,
which is used to derive the formula of marginal probabilities for \CRF{} and \BCRF{}.
Although this can be proved via Danskin's theorem \cite{danskin1966,bertsekas1997nonlinear}, we give here a simple alternate proof.

Let $h: \R^d \to \R \cup \{\infty\}$ be a function.
We first note that the Fenchel conjugate of $h$ is convex as it is the maximum of a set of affine functions, no matter if $f$ is convex or not.

\begin{thm}
  Let $h: \R^k \to \R \cup \{\infty\} $ be a function and $\vv \in \dom h^*$.
  Then, the following formula can be used to build a subgradient of $h^*$ at $\vv$:
  \[
    \partial h^*(\vv)
    \supseteq
    \argmax_{\vt \in \dom h}~\langle \vt, \vv \rangle - h(\vt).
  \]
  Moreover, if $h^*$ is differentiable at $\vv$, $\partial h^*$ is a singleton \cite[Th.\ 3.33]{beck2017firstorder}.
\end{thm}

\begin{proof}
Let $\widehat \vt$ be defined as follows:
\[
\widehat\vt \in \argmax_{\vt \in \dom h}~\langle \vt, \vv \rangle - h(\vt).
\]
We have $\widehat\vt \in \partial h^*(\vv)$ if and only the subgradient inequality holds \cite[Def.\ 3.1]{beck2017firstorder}, that is:
\[
    \forall \vv' \in \dom h^*:
    h^*(\vv') \geq h^*(\vv) + \langle \widehat\vt, \vv' - \vv\rangle.
\]
Starting from the right-hand side, for all $\vv' \in \dom h^*$, we can write:
\begin{align*}
    &h^*(\vv) + \langle\widehat \vt, \vv' - \vv\rangle
    \\
    &= \cancel{\langle\widehat \vt, \vv\rangle} - h(\widehat \vt) + \langle\widehat \vt, \vv'\rangle - \cancel{\langle\widehat \vt, \vv\rangle} \\
\intertext{where we simply replaced $h^*(\vv)$ by its definition using the fact that $\widehat \vt$ is a solution of the maximization problem. We derive an upper bound on this formula by maximizing over possible values for $\vt$:}
    &= \langle\widehat \vt, \vv'\rangle  - h(\widehat \vt) \\
    &\leq \max_{\vt \in \dom h} \langle\vt, \vv'\rangle  - h(\vt) \\
    &= h^*(\vv')
\end{align*}
Hence the subgradient inequality holds, and $\widehat \vt$ is a subgradient of $h^*$ at $\vv$.
\end{proof}

%% file: 9.proof_entropic_reg.tex
\section{Proof of Proposition \ref{th:limit}}
\label{app:proof_limit}

Linear programming with entropic regularization is a well-studied setting \cite{Fang1992,Fang1997}.
Nonetheless, we adapt the proof of \cite[][Prop.\ 4.1]{peyre2019compot} to our problem for completness.

\begin{proof}
In this proof, we will write:
\begin{align*}
    \widehat \vy &\in \argmax_{\vy \in \conv(Y)}~\langle \vw, \vy\rangle
    \\
    \text{and}\quad\widehat \vmu^{(\tau)} &= \argmax_{\vy \in \conv(Y)}~\langle \vw, \vy\rangle + \tau H(\vy)
    \\
    &= \nabla B_Y(\tau^{-1}\vw).
\end{align*}
By optimality of $\widehat \vy$, we have:
\begin{align*}
&\langle \vw, \widehat\vy\rangle \geq \langle \vw, \widehat\vmu^{(\tau)}\rangle
\\
\Longleftrightarrow\quad
&0 \leq \langle \vw, \widehat\vy\rangle - \langle \vw, \widehat\vmu^{(\tau)}\rangle.
\end{align*}
Similarly, by optimality of $\widehat \vmu^{(\tau)}$:
\begin{align*}
&\langle \vw, \widehat\vmu^{(\tau)}\rangle + \tau H(\widehat\vmu^{(\tau)}) \geq \langle \vw, \widehat\vy\rangle + \tau H(\widehat\vy)
\\
\Longleftrightarrow\quad
&\langle \vw, \widehat\vy\rangle - \langle \vw, \widehat\vmu^{(\tau)}\rangle \leq \tau (H(\widehat\vmu^{(\tau)}) - H(\widehat\vy))
\\
\end{align*}
Combining the two inequalities, we have:
\[
0 \leq \langle \vw, \widehat\vy\rangle - \langle \vw, \widehat\vmu^{(\tau)}\rangle
\leq \tau \underbrace{(H(\widehat\vmu^{(\tau)}) - H(\widehat\vy))}_{\geq 0}
\]
Notice that for any $\tau > 0$, we have $H(\widehat\vmu^{(\tau)}) \leq c$ where $c > 0$ is a constant.
Therefore:
$$
\lim_{\tau \to 0} \tau (H(\widehat\vmu^{(\tau)}) - H(\widehat\vy)) = 0.
$$
We can therefore apply the squeeze theorem:
\begin{align*}
    &\lim_{\tau \to 0} \left( \langle \vw, \widehat\vy\rangle - \langle \vw, \widehat\vmu^{(\tau)}\rangle\right) = 0
    \\
    \Longleftrightarrow
    \quad&
    \lim_{\tau \to 0} \langle \vw, \widehat\vmu^{(\tau)}\rangle =  \langle \vw, \widehat\vy\rangle,
\end{align*}
which ends the proof.
\end{proof}

%% file: 9.proof_polytope.tex
\section{Proof of Proposition \ref{thm:integrality}}
\label{app:proof_marginal_polytope}

\begin{proof}
Denote by $\mathcal{P}$ the polytope on the right-hand side of the theorem statement. Since there is a one-to-one correspondence between valid paths and binary points of $\mathcal{P}$, it remains to prove that $\mathcal{P}$ is integral.

Consider the directed graph $G'=(V',E')$ obtained from $G$ by adding two nodes $s$ and $t$, an arc $(s,\rv)$ for all $\rv \in V_1$, an arc $(\rv, t)$ for $\rv \in V_n$, and the arc $(t,s)$. 
By construction, every point \(\overline \vy\) of \(\mathcal{P}\) can be extended to a point of the following polytope:
\[
\mathcal{Q}
=
\left\{
    y \in \mathbb{R}^{E'}_+
\middle|
    \begin{array}{l}
         y_{ts} = 1,\\
         \forall v \in V':
         \begin{array}[t]{l}
            \sum_{a \in \delta^+(\rv)} y_a \\
            = \sum_{a \in \delta^-(\rv)} y_a
        \end{array}
    \end{array}
\right\}
\]
by setting \(\overline y_{(t,s)} = 1\), \(\overline y_{(s,\rv)} = \sum_{a \in \delta^+(\rv)} \overline y_a\) for all \(\rv \in V_1\), and \(\overline y_{(\rv,t)} = \sum_{a \in \delta^-(\rv)} \overline y_a\) for all \(\rv \in~V_n\). Hence, \(\mathcal{Q}\) is an extended formulation of $\mathcal{P}$ \citep{conforti_extended_2013}. Since $\mathcal{Q}$ is integral \citep[p.\ 274]{Schrijver1986}, so is~$\mathcal{P}$.
\end{proof}

%% file: 9.lightspeed.tex
\section{Iterative Bregman Projections}\label{app:IBP}

Recall that \(\conv(Y) = \mathcal{C}_{\text{even}} \cap \mathcal{C}_{\text{odd}}\) and solving the projections on \(\mathcal{C}_{\text{even}}\) and \(\mathcal{C}_{\text{odd}}\) decompose into solving several problems of the form:
\[
    \min_{\vq\vert_{\delta(V_i)} \in \mathcal{C}_{i}} \KL\left(\vq\vert_{\delta(V_i)},  \vw\vert_{\delta(V_i)} \right),
\]
for some \(i \in \llbracket 2,n-1 \rrbracket\).
This latter problem can be reformulated as:
\begin{align}
	\max_{\vq}~& \sum_{a \in \delta(V_i)} \big( (\evw_a + 1) \evq_a  - \evq_a \log \evq_a \big) \label{sysSPBcluster:obj} \\
	\text{s.t.}~& \sum_{a \in \delta^-(\rv)} \evq_a = \sum_{a \in \delta^+(\rv)} \evq_a, \quad \forall \rv \in V_i, \label{cst:subpb:FC}\\
	&\sum_{a \in \delta^-(V_i)} \evq_a = 1, \label{cst:subpb:EC}\\
        & \vq \ge \bm{0} \label{sysSPBcluster:nonNeg}
\end{align}
Therefore, in the following we show that this problem has a closed-form expression.

Dualizing Constraints \eqref{cst:subpb:FC} and \eqref{cst:subpb:EC} gives the following Lagrangian function:\footnote{For a sake of simplicity, we ignore inequalities \( \vq \ge \bm{0}\) as by \eqref{eq:valueYenteringArc} and \eqref{eq:valueYleavingArc}, \(\vq\) is nonnegative for all \(\vlambda, \nu\), following similar argument to the one in Appendix~\ref{app:logsumexp}.}
\begin{align*}
\mathcal L(\vq, \vlambda, \nu)
=&
\sum_{a \in \delta(V_i)} \big((\evw_a + 1)\evq_a  - \evq_a \log \evq_a \big)\\
&+ \sum_{\rv \in V_i} \lambda_v \left(\sum_{a \in \delta^-(\rv)} \evq_a - \sum_{a \in \delta^+(\rv)} \evq_a \right)\\
&+ \nu\left(\sum_{a \in \delta^-(V_i)} \evq_a - 1\right).
\end{align*}
For fixed \(\vlambda\) and \(\nu\), the associated Lagrangian relaxed problem \(\mathcal{L}(\vlambda, \nu)\) is: 
\begin{equation}
    \mathcal{L}(\vlambda, \nu) = \max_{\vq} \mathcal L(\vq, \vlambda, \nu),
\end{equation}
and the Lagrangian dual problem \(\mathcal{L}\) is:
\begin{equation}
    \mathcal{L} = \min_{\vlambda, \nu} \mathcal{L}(\vlambda, \nu).
\end{equation}
Since \eqref{sysSPBcluster:obj} is concave, strong duality holds, that is, the optimum of \eqref{sysSPBcluster:obj}-\eqref{sysSPBcluster:nonNeg} equals the one of \(\mathcal{L}\). Moreover, an optimal solution \(\widehat \vq\) of \eqref{sysSPBcluster:obj}-\eqref{sysSPBcluster:nonNeg} and an optimal \(\widehat\vlambda, \widehat\nu\) of \(\mathcal{L}\) satisfy the KKT conditions \cite[Sec.\ 5.5.3]{boyd2004convex}.

The stationarity condition for $a \in \delta^-(\rv)$ implies that:
\begin{align}
    \nonumber
    \hspace{-0.8em}\frac{\partial}{\partial \evq_a} \mathcal{L}(\widehat\vq, \widehat\vlambda, \widehat\nu)
    &=
    0
    \\
    \nonumber
    \log \widehat\evq_a
    &=
    \evw_a + \widehat\lambda_\rv + \widehat\nu
    \\
	\widehat\evq_a
    &= \exp(\evw_a)\exp(\widehat\lambda_\rv)\exp(\widehat\nu).
    \label{eq:valueYenteringArc}
\end{align}
Similarly, the stationarity condition for $a \in \delta^+(\rv)$ implies that:
\begin{align}
\nonumber	\frac{\partial}{\partial \evq_a} \mathcal{L}(\widehat\vq, \widehat\vlambda, \widehat\nu) &= 0 \\
\nonumber	\log \widehat\evq_a &= \evw_a - \widehat\lambda_\rv \\
	\widehat\evq_a &= \frac{\exp(\evw_a)}{\exp(\widehat\lambda_\rv)} \label{eq:valueYleavingArc}
\end{align}
At optimality, \(\widehat \vq\) is primal feasible. Hence, it satisfies \eqref{cst:subpb:FC} so we have for all $\rv \in V_i$:
\begin{equation*}
    	\sum_{a \in \delta^-(\rv)}
        \hspace{-0.5em}
        \exp(\evw_a)\exp(\widehat\lambda_\rv)\exp(\widehat\nu) =
        \hspace{-0.5em}
        \sum_{a \in \delta^+(\rv)} \frac{\exp(\evw_a)}{\exp(\widehat\lambda_\rv)},
\end{equation*}
which gives:
\begin{equation}
	\widehat\lambda_\rv = \frac{1}{2}\log w^+(\rv) -  \frac{1}{2}\log w^-(\rv) -  \frac{1}{2} \widehat\nu,    \label{eq:valueLambdaV}
\end{equation}
where:
\begin{align*}
    w^+(\rv) &= \sum_{a \in \delta^+(\rv)} \exp(\evw_a) \\
    \text{and~}
    w^-(\rv) &= \sum_{a \in \delta^-(\rv)} \exp(\evw_a).
\end{align*}
Since \(\widehat\vq\) satisfies~\eqref{cst:subpb:EC}, we have
\begin{equation*}
    \sum_{\rv \in V_i} \sum_{a \in \delta^-(\rv)} \exp(\evw_a)\exp(\widehat\lambda_\rv)\exp(\widehat\nu) = 1.
\end{equation*}
By \eqref{eq:valueLambdaV}, we get:
\begin{equation*}
\exp\left(-\frac{1}{2}\widehat\nu\right) = \sum_{\rv \in V_i} w^-(\rv) \exp\big(  \sigma(\rv)\big)   
\end{equation*}
where:
\[
    \sigma(\rv) = \frac{1}{2} \log w^+(\rv) - \frac{1}{2} \log w^-(\rv).
\]
for all \(\rv \in V_i\).
This gives:
\begin{equation}
    \widehat\nu = -2 \log \left( \sum_{\rv \in V_i}  w^-(\rv) \exp(\sigma(\rv))\right). \label{eq:valueMuOpt}
\end{equation}
Finally, replacing \(\widehat\vlambda\) and \(\widehat\nu\) by their value given by \eqref{eq:valueLambdaV} and \eqref{eq:valueMuOpt} in the formulas \eqref{eq:valueYenteringArc} and \eqref{eq:valueYleavingArc} gives:
\begin{align*}
    &\widehat \evq_{(u,u')}
    \\
    &= \left\{ \begin{array}{ll} 
    \frac{\exp(w_{(u,u')})}{\exp(\sigma(u))\left(\sum_{\rv \in V_i} w^-(\rv) \exp(\sigma(\rv))\right)} & \text{if \(u \in V_i\),}\\
    \frac{\exp(w_{(u,u')})\exp(\sigma(u'))}{\sum_{\rv \in V_i} w^-(\rv) \exp(\sigma(\rv))} & \text{if \(u' \in V_i\),}
    \end{array} \right.
\end{align*}
for all $(u,u') \in \delta(V_i)$.

%% file: 9.mft.tex
\section{Mean Field Theory}\label{sec:mean-field-theory}
Mean field (\MF{}) approximation for \CRF{}~\cite{wainwright2008expfam,zheng-2015-condit-random,wang-etal-2020-ain} is a general method which aims at approximating a \MRF{} with a tractable distribution, in the case of tagging approximate a linear-chain \CRF{} with a factorized distribution (the so-called \emph{naive} \MF{}) of the form:
\[
    r(\vx|\vs) = \prod_{i} r_{i}(\evx_{i}| \vs).
\]
The main idea is to search for the factorized distribution that is the closest to the
\CRF{} distribution $p(\cdot|\vs)$ in terms of their Kullback-Leibler
divergence~\cite[Chap.~5.2.2]{wainwright2008expfam}:
\[
  \argmin_{r} D_{KL}(r(\cdot|\vs), p(\cdot|\vs)).
\]
Even if it's not necessarily obvious at first glance, the key advantage of this approach is that the log-partition term $A_Y$ appears as a constant in the objective, hence there is no need to rely on the forward algorithm to search for the optimal \MF{} distribution.

\MF{} inference is often implemented using iterative parallel updates.
Starting from a any distribution $r^{0}$, each iteration computes an updated distribution as follows:
\[
r_{i}^{k}(t) \leftarrow \frac{\exp \big(m(i,t,k) \big)}{\sum_{t'} \exp \big(m(i,t',k)\big)  },
\]
where variables $m$ are ``messages'' incoming from adjacent tag distributions defined as follows::
\begin{align*}
 m_{\rightarrow}(i,t,k) &= \mathbb{E}_{t'\sim r_{i-1}^{k-1}(\cdot)}[f_{\vtheta}(\vs)_{i-1,t',t}] \\
 m_{\leftarrow}(i,t,k) &= \mathbb{E}_{t'\sim r_{i+1}^{k-1}(\cdot)}[f_{\vtheta}(\vs)_{i,t,t'}] \\
 m(i,t,k) &= m_{\rightarrow}(i,t,k) + m_{\leftarrow}(i,t,k).
\end{align*}
Since the \MF{} inference objective is non-convex, the quality of this method relies on the initial distribution.
Moreover, the parallel update procedure is not guaranteed to converge
\cite{kraehenbuehl13mftconvergence}.

Note that given the update formulas, \MF{} cannot take into account well-formedness constraints, \ie{} we cannot forbid adjacent pair of tags by removing transitions or setting transition weights to $-\infty$.

%% file: 9.loss.tex
\section{Partial Fenchel-Young Losses}
\label{app:loss}

In this appendix, we prove several properties of partial FY losses that motivate their use for learning from partial labels.

\paragraph{Generalization of FY losses.}
If $\widetilde Y\!=\!\{ \vy \}$ is a singleton containing only a gold label, then
\(
\widetilde{\ell}_{\Omega}(\vw, \{ \vy \}) = \ell_\Omega(\vw, \vy).
\)

\begin{proof}
When $\widetilde Y = \{\vy\}$ is a singleton, we have $\conv(\widetilde Y) = \{ \vy \}$ and therefore:
\begin{align*}
    &\hspace{-1em}\widetilde{\ell}_{\Omega}(\vw, \{\vy\})\\
    =&
    \left(\Omega + \delta_{\conv(Y)}\right)^*(\vw)
    - \left(\Omega + \delta_{\{ \vy \}}\right)^*(\vw)
    \\
    =&
    \left(\Omega + \delta_{\conv(Y)}\right)^*(\vw) \\
    &
    - \left( \sup_{\vmu \in \dom \Omega}~\langle \vmu, \vw \rangle - \Omega(\vmu) - \delta_{\{ \vy \}}(\vmu) \right) .\\
\intertext{Note that the indicator function restrict the search space of the maximization to a single element $\vy$, therefore:}
    &= \Omega^*(\vw) - \langle \vmu, \vy \rangle + \Omega(\vy) \\
    &= \ell_\Omega(\vw, y),
\end{align*}
which ends the proof.
\end{proof}

\paragraph{Non-negativity.}
The loss is bounded below by 0.

\begin{proof}
Note that we have:
\begin{align*}
    &\left(\Omega + \delta_{\conv(\widetilde Y)}\right)^*(\vw)
    \\
    &= \sup_{\vmu \in \dom \Omega} \langle \vmu, \vw \rangle - \Omega(\vmu) - \delta_{\conv(\widetilde Y)}(\vmu). \\
\intertext{The indicator function act as a constraint on $\vmu$. By definition we have $\widetilde Y \subset Y$, and therefore by increasing the search space we derive an upper bound:}
    &\leq \sup_{\vmu \in \dom \Omega} \langle \vmu, \vw \rangle - \Omega(\vmu) - \delta_{\conv(Y)}(\vmu) \\
    &= \left(\Omega + \delta_{\conv(Y)}\right)^*(\vw).
\end{align*}
Therefore, the loss is non-negative.
\end{proof}

\paragraph{Smaller partial labeling set $\implies$ bigger loss.}
Let $\widetilde Y' \subseteq \widetilde Y$, then:
\[
\widetilde{\ell}(\vw ; \widetilde Y') \geq \widetilde{\ell}(\vw ; \widetilde Y).
\]

\begin{proof}
Note that if $\widetilde Y' \subseteq \widetilde Y$, then $\conv(\widetilde Y') \subseteq \conv(\widetilde Y)$.
By definition, we have:
\begin{align*}
    &\hspace{-1em}\widetilde{\ell}_{\Omega}(\vw ; \widetilde Y')
    \\
    =&
    \left(\Omega + \delta_{\conv(Y}\right)^*(\vw)
    - \left(\Omega + \delta_{\conv(\widetilde Y')}\right)^*(\vw)
    \\
    =&
    \left(\Omega + \delta_{\conv(Y)}\right)^*(\vw) \\
    &
    - \left( \sup_{\vmu \in \dom \Omega}~\langle \vmu, \vw \rangle - \Omega(\vmu) - \delta_{\conv(\widetilde Y')}(\vmu) \right) .\\
\intertext{If we maximize over a larger set, the second term will be larger therefore:}
    \geq&\left(\Omega + \delta_{\conv(Y)}\right)^*(\vw)
    \\
    &- \left( \sup_{\vmu \in \dom \Omega} \langle \vmu, \vw \rangle - \Omega(\vmu) - \delta_{\conv(\widetilde Y)}(\vmu) \right) \\
    =&\widetilde{\ell}_{\Omega}(\vw ; \widetilde Y),
\end{align*}
which ends the proof.
\end{proof}

%% file: 9.exp.tex
\section{Neural Network Hyperparameters}
\label{sec:train-impl}

We use standard self-attentive networks with embedings of size 768, 8 heads per layers and hidden dimension projection of 2048.
The model are training using the Adam optimizer \cite{adam} with a learning rate of $3 \times 10^{-4}$ when the transformer is learned from scratch, and $3 \times 10^{-5}$ when fine-tuning \textsc{BERT}.
We use a linear rate scheduler with warmup on $10\%$ of updates.

For part-of-speech tagging, we sum character-level embedding to word-level embedding at the input of the transformer, where the character level embeddings are obtained using a simple 1D convolution layer.